\documentclass{article}

\usepackage[final, nonatbib]{neurips_2023}

\usepackage[utf8]{inputenc} % allow utf-8 input
\usepackage[T1]{fontenc}    % use 8-bit T1 fonts
\usepackage{hyperref}       % hyperlinks
\usepackage{url}            % simple URL typesetting
\usepackage{booktabs}       % professional-quality tables
\usepackage{amsfonts}       % blackboard math symbols
\usepackage{nicefrac}       % compact symbols for 1/2, etc.
\usepackage{microtype}      % microtypography
\usepackage{xcolor}         % colors

\usepackage{algorithm}
\usepackage[noend]{algpseudocode}

\usepackage{amsmath}
\usepackage{amsthm}
\usepackage{amssymb}
\usepackage{bbold}
\usepackage{tikz}
\usetikzlibrary{calc} 
\usepackage{float}

\usepackage{xcolor}  % Add the xcolor package

\newtheorem{theorem}{Theorem}[section]

\title{Naturally Private Recommendations with Determinantal Point Processes}

\author{%
  Jack Fitzsimons \\
  Oblivious \\
  \texttt{jack@oblivious.com} \\
  \And
  Agustín Freitas Pasqualini \\
  Pento \\
  \texttt{agustin.freitas@pento.ai} \\
  \And
  Robert Pisarczyk \\
  Oblivious \\
  \texttt{rob@oblivious.com} \\
    \And
  Dmitrii Usynin \\
  Technical University of Munich \\
  \texttt{dmitrii.usynin@tum.de} \\
}

\begin{document}

% Notes:
% - Need to address the sampling based on eigenvector too  [To Do]
% - Text is very dense, try to make it more readable [To Do]
% - Used $L_2$ sensitivity but exp is $L_1$ based, so would need to multiply by sqrt(n) [Done]
% - May want to add some visual representation to engage the reader [To Do]
% - Proof read for typos and grammar [To Do]
\maketitle

\begin{abstract}
  Often we consider machine learning models or statistical analysis methods which we endeavour to alter, by introducing a randomized mechanism, to make the model conform to a differential privacy constraint. However, certain models can often be implicitly differentially private or require significantly fewer alterations. In this work, we discuss Determinantal Point Processes (DPPs) which are dispersion models that balance recommendations based on both the popularity and the diversity of the content. We introduce DPPs, derive and discuss the alternations required for them to satisfy $\epsilon$-Differential Privacy and provide an analysis of their sensitivity. We conclude by proposing simple alternatives to DPPs which would make them more efficient with respect to their privacy-utility trade-off.  
\end{abstract}

\section{Introduction}

%\textcolor{red}{TODO: Should probably say why these are important in RS}

%In machine learning and statistical analysis, the integration of privacy-preserving mechanisms into traditional models has become increasingly important due to growing data privacy concerns.
%Among the many models of interest to the community, Determinantal Point Processes (DPPs) \cite{kulesza2012determinantal} stand out as not only popular in recommendation systems but also inherently aligned with differential privacy principles.
%This work explores the intrinsic properties of DPPs that lend themselves to $\epsilon$-Differential Privacy and proposes methodologies to enhance their efficiency and privacy-utility trade-off in real-world applications.

Determinantal Point Processes, introduced by \cite{macchi1975coincidence}, arose in the study of fermionic gases in physics. They are a point process which samples sets of points by balancing the samples' prominence and the diversity of the sample set. Due to their statistical properties \cite{scardicchio2009statistical} and appropriateness as diversity-inducing priors \cite{kulesza2012determinantal}, they have become popular within the machine learning community \cite{gardner2018gpytorch, hennig2016exact}, recommender systems \cite{wilhelm2018practical, chen2018fast, gan2020enhancing, liu2020diversified, wu2019recent} and in probablistic numerics \cite{bardenet2020monte}.

While not intentionally used as a privacy-preserving mechanism, its form and stochasticity have a natural alignment with the exponential mechanisms widely leveraged as a mechanism to achieve differential privacy. This work aims to explore this relationship and present the conditions for the process to have a bounded privacy loss. 

\section{Background}

\subsection{Determinantal Point Processes (DPPs) and L-ensembles}

For machine learning, a particularly useful class of DPPs is known as \emph{L-ensembles} defined by a probability measure on subsets \(Y \subseteq \mathcal{Y}\) such that:

\[
P_L(Y) \propto \det(\mathbf{L}_Y)
\]

where \(\mathbf{L}_Y\) is the submatrix of \(\mathbf{L}\), a real symmetric positive definite matrix indexed by elements in \(\mathcal{Y}\), and \(\mathbf{L}\) is a kernel matrix which encodes the similarities between items. The normalization constant is \(\det(\mathbf{L} + \mathbf{I})\), with \(\mathbf{I}\) as the identity matrix.

\subsection{Sampling from a DPP Using Eigenvalue Decomposition}

To sample from a DPP, we first perform an eigenvalue decomposition of the kernel matrix \(\mathbf{L}\). Let \(\mathbf{L} = \mathbf{V}\mathbf{\Lambda}\mathbf{V}^T\), where \(\mathbf{V}\) is the matrix of eigenvectors and \(\mathbf{\Lambda}\) is the diagonal matrix of eigenvalues. The marginal kernel \(\mathbf{K}\) is then expressed as:

\[
\mathbf{K} = \mathbf{V}\mathbf{\Lambda}_K\mathbf{V}^T \quad \text{where} \quad \mathbf{\Lambda}_K = \text{diag}\left(\frac{\lambda_1}{\lambda_1 + 1}, \dots, \frac{\lambda_N}{\lambda_N + 1}\right)
\]

The sampling algorithm proceeds in two phases: eigenvector selection and subset construction, as outlined below.

\begin{algorithm}
\caption{Sampling from a DPP using Eigenvalue Decomposition}
\begin{algorithmic}[1]
    \State \textbf{Input:} Eigenvalue decomposition $\mathbf{L} = \mathbf{V}\mathbf{\Lambda}\mathbf{V}^T$ \Comment{Decomposition of $\mathbf{L}$}
    \State \textbf{Output:} Subset $Y$ sampled from the DPP

    \Procedure{EigenvalueSampling}{$\mathbf{\Lambda}, \mathbf{V}$}
        \State $V \gets \{v_i \mid \text{CoinFlip}(\frac{1}{\lambda_i + 1}, \frac{\lambda_i}{\lambda_i + 1}) \}$ \Comment{Select vectors using Weighted Coin Flip}
        \State \Return $V$ 
    \EndProcedure

    \Procedure{SubsetConstruction}{$V, \mathcal{Y}$}
        \State $Y \gets \emptyset$, $V' \gets \text{orthonormal basis of } V$ \Comment{Initialize $Y$ and $V'$}
        \While{$V' \neq \emptyset$}
            \State $i \gets \text{randomly from } \mathcal{Y} \text{ with weighted probability} \|\text{Proj}_{V'}(e_i)\|^2$ \Comment{Select an item $i$}
            \State $Y \gets Y \cup \{i\}$, $\mathcal{Y} \gets \mathcal{Y} \setminus \{i\}$ \Comment{Accept item $i$}
            \State Update $V'$ to be orthogonal to $e_i$ \Comment{Ensure sample without replacement}
        \EndWhile
        \State \Return $Y$ 
    \EndProcedure

    \State $V \gets \Call{EigenvalueSampling}{\mathbf{\Lambda}, \mathbf{V}}$ \Comment{Sample vectors}
    \State $Y \gets \Call{SubsetConstruction}{V, \mathcal{Y}}$ \Comment{Construct subset}
\end{algorithmic}
\end{algorithm}

This sampling method ensures that the selected subset \(Y\) is diverse by leveraging the orthogonality of the eigenbasis, reflecting the repulsive nature inherent in the DPP defined by \(\mathbf{L}\). The algorithm leverages the eigendecomposition of \(\mathbf{L}\) to efficiently handle the probabilistic selection and orthogonal projection steps, crucial for maintaining the diversity of the sampled subset.

\subsection{Kernel Construction for DPPs}
The kernel matrix $\mathbf{L}$ in DPPs, which encodes item similarities determines item diversity. Commonly, $\mathbf{L}$ is constructed using dot product kernels from count vectors, where $L_{ij} = \mathbf{x}_i \cdot \mathbf{x}_j$ for articles $i$ and $j$, ensuring $\mathbf{L}$ is positive semidefinite.

Alternative methods, including standard kernel functions and neural networks, also contribute to constructing $\mathbf{L}$. For instance, radial basis function (RBF), polynomial, and sigmoid kernels offer alternative similarity measures:
\[
L(x, y) =
\left\{
\begin{array}{ll}
e^{-\gamma \|x-y\|^2} & \text{(RBF Kernel)}, \\
(x \cdot y + c)^d & \text{(Polynomial Kernel)}, \\
\tanh(\alpha x \cdot y + c) & \text{(Sigmoid Kernel)}
\end{array}
\right.
\]

Pre-trained neural networks can be used to embed data into a Hilbert space $\phi(\mathbf{x}_i)$, $\phi(\mathbf{x}_j)$ for items, used in kernel functions $\phi(\mathbf{x}_i)\phi(\mathbf{x}_j)$ to produce sophisticated representations in $\mathbf{L}$, accommodating diverse data types and enhancing model applicability in domains such as computer vision, audio, and multimodal domains.

For the purpose of this work, we focus on sampling from DPPs given a known embedding, as opposed to model or hyperparameter selection. 

\section{Privacy Analysis of DPPs}

The core of this work is in the identification of the sampler used (sampling basis based on normalised eigenvalues) in the DPP context and its relationship to a vector response from the exponential mechanism,

\[
\mathbf{P}(\{v_i\} \in V) =  \frac{\lambda_i}{\lambda_i + 1},
\]

and its relationship to the vector response produced by the exponential mechanism,

\[
\mathbf{P}(x_i) =  \frac{e^{\frac{\epsilon u(x)}{2\Delta_u}} }{e^{\frac{\epsilon u(x)}{2 \Delta_u}} + 1}.
\]

It can easily be seen that the sampling technique is equivalent, assuming a constant sensitivity $\Delta u$, when the scoring function is set to  $\log{\lambda_i}$.

As a large number of kernels can be used to construct $L$, with varying sensitivity, let us define $\Delta_L \geq \|L - \Tilde{L}\|_2$ where $\Tilde{L}$ differs from $L$ by one subject. We will perform the sensitivity analysis in terms of $\Delta_L$.

Further, once we have sampled the eigenvalues, we can perform a similar interpretation of the eigenbasis sampling. The total privacy budget spent is a combination of both of these.

\textbf{Special Case:} A common scenario of the above is simply when $\phi(x) \in \mathbb{R}^{n \times m}$ represents a binary matrix indicating whether subject $i$ read news item $j$. The resulting kernel matrix $L = \phi(x)\phi(x)^T \in \mathbb{R}^{n \times n}$ indicates how popular each news item was and how many readers have read each pair-wise combination of articles. This is useful for our intuition as the sensitivity reduces to $\Delta_L = \|\mathbb{1}^{n \times n}\|_2 = n$, with $n$ equal to the number of items which can be selected from and $\mathbb{1}$ denotes a ones matrix. The number of items to select from is assumed to be a known, fixed quantity.

\subsection{Sensitivity Analysis of DPPs}

\subsubsection{Eigenvalue Sampling (with Preconditioning)}

To investigate the sensitivity of the DPP, we need to look at the scoring function $u(\lambda_i) = \log(\lambda_i)$. As this is a vector response, we can consider $\| u(\lambda_i) \forall i \in [1, n]\|_2^2 = \sum_i \log(\lambda_i)^2$. As, $\Delta_L^2 = \sum_i \lambda_i^2$, we can reason about the maximum change in $u$ given a shift of mass $\sqrt{n} \Delta_L$ over the eigenvalue vector $\lambda$\footnote{Note the scaling factor of $\sqrt{n}$ to accomodate the switch from $L_2$ to $L_1$ sensitivity due to the following vector norm inequality: $\|x\|_1\leq \sqrt{n}\| x\|_2$ for all $x\in \mathbb{R}^n$.}. Further, as the $\log(\lambda_i)$ function is monotonic with respect to $\lambda_i$, with gradient equal to $\frac{1}{\lambda_i}$, and as kernel matrices are positive semidefinite ($\lambda_i \geq 0$), the change of scoring function is maximised when $\lambda_i \rightarrow 0$. If any $\lambda_i = 0$, $\Delta_u = \infty$ and differential privacy is not achieved.

\begin{theorem}
An L-ensemble-based Determinantal Point Process (DPP) with kernel matrix $\mathbf{L}$ achieves $\epsilon$-differential privacy only if $\mathbf{L}$ is strictly positive definite, thereby preventing any eigenvalue $\lambda_i$ from approaching zero and ensuring the defined scoring function $u(\lambda) = \log(\lambda)$ remains finite.
\end{theorem}

In many kernel method techniques, similar restrictions on the condition of the matrix are required. Probably the most popular approach to dealing with this is the introduction of "jitter" \cite{gardner2018gpytorch, bernardo1998regression}, that is the addition of $\mathbf{I}\sigma$ which in turn bounds all eigenvalues by $\sigma$. We will assume the kernel matrix has a lower bound on eigenvalues of $\sigma$ for the rest of the analysis.

The most extreme scenarios of eigenspectra are when the eigenvalues are uniformly close to the magnitude of the jitter added. As the logarithm function has a monotonic gradient, perturbations on eigenvalues closest to zero will have the largest effect on $u$. Thus the most extreme change in $u$ with bounds $\Delta_u$ is when all eigenvalues change from $\sigma$ to $\sigma + \frac{\Delta_L}{\sqrt{n}}$.

This can be shown simply as the increase in $\Delta_u$ is greatest when the change in $|\frac{\partial \log(\lambda)}{ \partial \lambda}|$ is greatest and hence evenly spreading the magnitude across the smallest eigenvalues, which are as close to zero as possible while maintaining the jitter constraint. The bound on the sensitivity can thus be set as $\Delta_u = n (\log(\sigma + \frac{\Delta_L}{\sqrt{n}}) - \log(\sigma)) = n \log \left( 1 +\frac{  \Delta_L}{\sigma \sqrt{n}} \right)$.

\subsubsection{Eigenvector Sampling (with Preconditioning)}
For the full sensitivity analysis of the DPP, we also need to consider the \texttt{SubsetConstruction} step of the sampling algorithm. We can again consider the scoring function to be logarithmic $u(e_i,V')=\log(\|\text{Proj}_{V'}(e_i)\|^2)$. Since $\|\text{Proj}_{V'}(e_i)\|^2$ can in principle be zero, so as before we add an appropriate jitter $\delta$ to matrix \(\mathbf{V}\) s.t. the probabilities are proportional to $\|\text{Proj}_{V'}(e_i)\|^2+\delta$. Since the gradient of $\log(x)$ is $\frac{1}{x}$, the biggest change in the scoring function given a shift of mass $s$ is when all but one components are as small as possible (given the jitter condition) and most of the initial mass is on a single component of the vector. Starting with $V'=V$, the components change as
\[
(1+n\delta)^{-1}\left[
    1+ \delta, \hspace{2mm} \delta, \hspace{2mm} \ldots, \hspace{2mm} \delta
\right]
\rightarrow
(1+n\delta+s)^{-1}\left[
    1+ \delta, \hspace{2mm} \delta+\frac{s}{n-1}, \hspace{2mm} \ldots, \hspace{2mm} \delta+\frac{s}{n-1}
\right]
\]
The total bound on the sensitivity can be simply upperbounded by $n$ instances of this change as $\Delta_u = n(n-1)\log\left(1+\frac{s}{(n-1)\delta}\right)-n\log\left(1+\frac{s}{1+n\delta}\right)$.
What remains is to relate $s$ i.e. perturbation in the eigenvectors to $\Delta_L$. Due to the Davis-Kahan theorem \cite{davis1970rotation}, the perturbation depends on the separation of the eigenvalues ("eigengap"). For well separated eigenvalues, the perturbation to eigenvectors is stable and small. As for many kernels e.g. the Gaussian kernels, eigenvalues decay exponentially, the largest eigenvalues i.e. the ones which are most likely to be sampled, are well-separated \cite{braun2005spectral}. However, we leave the formal analysis of this for future work. 
\subsection{Implicit Privacy Budget Consumption}

Considering that the \texttt{EigenvalueSampling} of the DPP algorithm naturally samples the eigenvalues with probability $\frac{\lambda_i}{1 + \lambda_i}$ and based on our interpretation of this being a vector response with the exponential mechanism, $\frac{\epsilon u(\lambda_i)}{2 \Delta_u} = \log(\lambda_i)$, we can easily determine $\epsilon$ for that step. As we set $u(\lambda_i) = \log(\lambda_i)$, we get the equality $\epsilon = 2 \Delta_u$. Plugging in our previous result from the sensitivity analysis we find the $\epsilon$ used in a regular DPP in the \texttt{EigenvalueSampling}:
$
\epsilon = 2 n \log \left( 1 +\frac{ \Delta_L}{\sigma \sqrt{n}} \right). 
$

However, in order to determine the total $\epsilon$ in terms of $\Delta_L$, we also need to include the $\epsilon$ for the \texttt{SubsetConstruction}. As discussed in the previous section, the dependence on  $\Delta_L$ needs to be quantified based on the eigenvector perturabion which we leave for future work.

\section{Shortcomings and Future Work}
As we have previously demonstrated, one can obtain natural privacy guarantee for the eigenvalue sampling in DPPs.
However, there exists a number of limitations that have not yet been discussed in-detail if one wanted to consciously use DPPs in a privacy-preserving setting:

\begin{enumerate}
    \item The sensitity analysis for \texttt{SubsetConstruction} step requires the consideration of the separation of the eigenvalues, which can be expected to be well-behaved for kernels with exponentially decaying eigenvalues but is challenging to bound in general. 
    \item On every recommendation more epsilon would be consumed due to the use of the exponential mechanism. 
    \item Depending on the kernel, jitter and number of items considered, the $\epsilon$ used may be prohibitive to provide an acceptable privacy guarantee. 
    \item Increased jitter will improve the stability and in turn the $\epsilon$ spent, but at a cost of potentially selecting more unpopular items than desired.
\end{enumerate}

As a result, we believe that consideration of DPP with an addition of a typical DP mechanism (instead of interpreting it as a stand alone privacy mechanism) could be beneficial to the scientific community and outline this as a promising avenue for future work. A natural choice would be to apply techniques such us the Sparse Vector Technique (SVT) \cite{hardt2010multiplicative} to both the \texttt{EigenvalueSampling} and \texttt{SubsetConstruction} steps.

\bibliographystyle{unsrt}
\bibliography{main}
\end{document}